\documentclass[11pt]{article}

\usepackage[final]{acl}

\usepackage{latexsym}
\usepackage{microtype}
\usepackage{graphicx}

\usepackage{fontspec}
\usepackage{polyglossia}

\setmainlanguage{english}
\setotherlanguage{arabic}

\usepackage{inconsolata}

\newfontfamily\arabicfont[Script=Arabic]{amiri-regular.ttf}

\usepackage[most]{tcolorbox}

\newtcolorbox{promptbox}{
  colback=gray!5,
  colframe=gray!40,
  arc=1.5mm,
  boxrule=0.6pt,
  left=10pt,       
  right=10pt,     
  top=8pt,        
  bottom=8pt,     
  fontupper=\small\ttfamily
}

\title{Rosetta at AlexandriaX-2026: LoRA-Adapted NileChat for Context-Aware Dialectal Arabic Dialogue Translation\thanks{Accepted to AlexandriaX 2026 Shared Task (Subtask 1) at ArabicNLP 2026.}}

\author{
  \textbf{Nada Esmaeil\textsuperscript{1}},
  \textbf{Fathima Rena\textsuperscript{2}},
  \textbf{Sibi Subhash\textsuperscript{2}},
  \textbf{Osama Elgendy\textsuperscript{3}},
  \\
  \textbf{Mina Naguib\textsuperscript{1}},
  \textbf{Salma Omar\textsuperscript{4}},
  \textbf{Muhammad Arif\textsuperscript{5}}
  \\[1.5ex]
  \textsuperscript{1}Tanta University, Egypt \quad
  \textsuperscript{2}Alliance University, India \quad
  \textsuperscript{3}Robusta Studio, Egypt \\
  \textsuperscript{4}Alexandria University, Egypt \quad
  \textsuperscript{5}Yale University, USA
  \\[1ex]
}

\begin{document}
\maketitle

\pagestyle{plain}
\thispagestyle{plain}

\begin{abstract}
This paper describes the Rosetta system for Subtask 1 (Context-Aware English-to-Dialectal Arabic Dialogue Translation) of the AlexandriaX shared task, participating in both constrained and unconstrained tracks. The approach fine-tunes a LoRA adapter on NileChat-3B using structured system/user prompts that condition generation on dialect and dialogue context. For the unconstrained track, the adapter is additionally pretrained on MADAR and PADIC. Rosetta ranked 4th in the constrained track (spBLEU 26.10) and 5th in the unconstrained track (spBLEU 25.09). The experimental results demonstrate that external pretraining helps only two of thirteen dialects while slightly hurting overall performance, suggesting negative transfer.
\end{abstract}

\section{Introduction}

Arabic exhibits strong diglossia \cite{ferguson-1959-diglossia}: Modern Standard Arabic (MSA) is the formal written variety, while regional dialects dominate everyday and online communication. Dialects diverge from MSA and from one another lexically, morphologically and pragmatically, and most lack standardized orthography or large parallel resources \cite{zaidan-callison-burch-2014-arabic}. Consequently, systems trained mainly on MSA generalize poorly to dialectal text.
 
AlexandriaX Subtask 1 \cite{elmekki-etal-2026-alexandriax} addresses this gap by requiring context-aware translation of English dialogue turns into a specified Arabic dialect. Systems receive the source turn together with preceding dialogue history, target country/dialect label, domain, speaker/addressee gender and (when available) persona information, and must produce only the translation of the current turn.
 
Participation spans both competition tracks under the official task constraints:
\begin{itemize}

    \item \textbf{Constrained}: only the provided Alexandria data; models $\le$ 5B parameters.

    \item \textbf{Unconstrained}: external data and larger models allowed.

\end{itemize}

The proposed approach fine-tunes a LoRA adapter \cite{hu-etal-2022-lora} on NileChat-3B \cite{el-mekki-etal-2025-nilechat}, an Arabic-pretrained instruction-tuned model, using structured prompts that encode dialect and conversational context. For the unconstrained track, the adapter is first pretrained on MADAR \cite{bouamor-etal-2018-madar} and PADIC \cite{meftouh-etal-2015-padic}. The system placed 4th (spBLEU \cite{goyal-etal-2022-flores} 26.10) and 5th (spBLEU 25.09) respectively. Performance is weakest on Mauritanian, Libyan, Sudanese and Moroccan varieties; external pretraining improves only Libyan and Moroccan while slightly degrading the overall average, and the pattern is not fully explained by dialect-family coverage alone. All code, fine-tuning scripts, and adapter weights are publicly available.\footnote{\url{https://github.com/NadaAdelMousa/Rosetta_at_AlexandriaX}}

\section{Background}

\subsection{Task setup} 

The input to Subtask 1 of the AlexandriaX shared task \cite{elmekki-etal-2026-alexandriax} is an English dialogue turn together with its
associated metadata (dialect/country label, domain, gender, dialogue
history, and persona where available); the output is the translation
of that turn only, in the target dialect, with prior turns supplied
as context rather than re-translated.

\subsection{Dataset} 
The organizers provide the Alexandria dataset \cite{el-mekki-etal-2026-alexandria} of dialectal Arabic dialogues spanning multiple countries/dialects. Table~\ref{tab:alexandria-splits} summarizes the number of turns per dialect across the train, development, and test splits.

\begin{table}[t]
\centering
\small
\setlength{\tabcolsep}{9pt}
\begin{tabular}{lrrr}
\hline
Dialect & Train & Dev & Test \\
\hline
Egyptian (EG)       & 3,108  & 1,113 & 1,113 \\
Jordanian (JO)      & 5,501  & 1,113 & 1,109 \\
Lebanese (LB)       & 8,906  & 1,118 & 1,110 \\
Libyan (LY)         & 0      & 0     & 1,309 \\
Moroccan (MA)       & 2,573  & 1,110 & 1,111 \\
Mauritanian (MR)    & 5,515  & 1,114 & 1,119 \\
Omani (OM)          & 6,280  & 1,109 & 1,107 \\
Palestinian (PS)    & 14,933 & 1,110 & 1,111 \\
Saudi (SA)          & 8,470  & 1,110 & 1,114 \\
Sudanese (SD)       & 0      & 0     & 915 \\
Syrian (SY)         & 6,071  & 1,119 & 1,114 \\
Tunisian (TN)       & 2,034  & 1,116 & 1,114 \\
Yemeni (YE)         & 3,089  & 1,118 & 1,113 \\
\hline
\textbf{Total}      & \textbf{66,480} & \textbf{12,250} & \textbf{14,459} \\
\hline
\end{tabular}
\caption{Number of turns in the Alexandria train, development, and test splits by dialect.}
\label{tab:alexandria-splits}
\end{table}

\subsection{Related work} 
Arabic's diglossic nature — a formal MSA standard alongside diverse spoken dialects — has long challenged machine translation, compounded by persistent data scarcity across dialects \cite{zbib-etal-2012-machine, sajjad-etal-2020-arabench, kadaoui-etal-2023-madar}. Early parallel resources remain limited in scale: PADIC \cite{meftouh-etal-2015-padic} covers a handful of Maghrebi and Levantine dialects, while MADAR \cite{bouamor-etal-2018-madar} offers translations into 25 city-level dialects within the travel domain. Such resources are typically narrow in domain and lack dialogue-level context \cite{malaysha-etal-2024-curating, taguchi-etal-2025-context}. The Alexandria dataset underlying this shared task addresses these gaps with a larger, conversation-based corpus spanning 13 Arab countries. In this work, we use MADAR and PADIC not as evaluation benchmarks but as external pretraining resources, aiming to leverage their broader dialectal coverage.

Adapting general-purpose LLMs to specific languages and cultures typically involves prompt engineering, culturally grounded fine-tuning, or continued pretraining on target-specific data \cite{bang-etal-2023-multitask, alkhamissi-etal-2024-investigating, naous-etal-2024-beer}. Arabic LLMs follow either path, trained from scratch \cite{sengupta-etal-2023-jais} or adapted from existing multilingual models \cite{bari-etal-2025-allam, fanar-team-2025-fanar}. NileChat \cite{el-mekki-etal-2025-nilechat} follows the latter approach, continuing pretraining of Qwen2.5-3B on synthetic and retrieval-based Arabic dialectal data, and outperforms other Arabic-aware models of comparable size while explicitly targeting specific dialectal communities (Egyptian and Moroccan Arabic) rather than treating Arabic as monolithic.

\section{System Overview}

The Rosetta system freezes UBC-NLP/NileChat-3B \cite{el-mekki-etal-2025-nilechat} and attaches a LoRA adapter ($r=32$, $\alpha=32$, dropout$=0$) \cite{hu-etal-2022-lora} targeting the attention and MLP projections (\texttt{q/k/v/o\_proj}, \texttt{gate/up/down\_proj}). Submissions were made to both the constrained and unconstrained tracks; the two configurations share the same architecture and differ only in the pretraining data used prior to task fine-tuning.

\subsection{Prompting Strategy}

Each example is presented as a system--user pair. Following the prompting guidelines established in the Alexandria dataset release \cite{el-mekki-etal-2026-alexandria}, the system prompt instructs the model to return only the translation while respecting meaning, tone and gender direction. The user prompt supplies the target dialect, metadata (country, domain, participants, speaker, gender direction) and the conversation history (speaker + source + translation for preceding turns). At inference, the model’s own previous outputs are fed back as history. A concrete example of this multi-turn conversational prompt is detailed in Appendix~\ref{sec:app-alexandria-prompt}.

\paragraph{Pretraining Prompt.}
Unlike the Alexandria prompts, the external pretraining prompts do not contain dialogue history or detailed conversational metadata. The system prompt instructs the model to act as a translator and to return only the translation, while the user prompt specifies the target dialect, the corresponding country, and the English sentence. An illustrative example of a MADAR instance prompt formatted under this scheme is provided in Appendix~\ref{sec:app-pretraining-prompt}.

\subsection{Tracks}
In the constrained track the adapter is fine-tuned solely on the Alexandria training set \cite{el-mekki-etal-2026-alexandria}. In the unconstrained track it is first pretrained on MADAR \cite{bouamor-etal-2018-madar} and PADIC \cite{meftouh-etal-2015-padic} (mapped to the same country-level dialect labels and converted to English--dialect pairs via NLLB-200 \cite{nllb-2022-scaling}), then fine-tuned on Alexandria data.

\section{Experimental Setup}

\subsection{Data Splits}

We use the official Alexandria splits \cite{el-mekki-etal-2026-alexandria} (Table~\ref{tab:alexandria-splits}). For the unconstrained track, MADAR \cite{bouamor-etal-2018-madar} and PADIC \cite{meftouh-etal-2015-padic} are additionally used during an adapter pre-training stage before fine-tuning on the Alexandria training data. The number of training and development examples used from MADAR and PADIC is summarized in Table~\ref{tab:madar-padic-statistics}.

\begin{table}[t]
\centering
\small
\setlength{\tabcolsep}{6pt}
\begin{tabular}{lrrrr}
\hline
& \multicolumn{2}{c}{\textbf{MADAR}} & \multicolumn{2}{c}{\textbf{PADIC}} \\
\textbf{Dialect} & Train & Dev & Train & Dev \\
\hline
Egyptian (EG)       & 13,800 & 1,600 & 0     & 0   \\
Jordanian (JO)      & 3,200  & 400   & 0     & 0   \\
Lebanese (LB)       & 10,600 & 1,200 & 0     & 0   \\
Libyan (LY)         & 3,200  & 400   & 0     & 0   \\
Moroccan (MA)       & 12,200 & 1,400 & 5,767 & 641 \\
Mauritanian (MR)    & 0      & 0     & 0     & 0   \\
Omani (OM)          & 1,600  & 200   & 0     & 0   \\
Palestinian (PS)    & 1,600  & 200   & 5,767 & 641 \\
Saudi (SA)          & 3,200  & 400   & 0     & 0   \\
Sudanese (SD)       & 1,600  & 200   & 0     & 0   \\
Syrian (SY)         & 3,200  & 400   & 5,767 & 641 \\
Tunisian (TN)       & 12,200 & 1,400 & 0     & 0   \\
Yemeni (YE)         & 1,600  & 200   & 0     & 0   \\
\hline
\textbf{Total}      & \textbf{68,000} & \textbf{8,000} &
\textbf{17,301} & \textbf{1,923} \\
\hline
\end{tabular}
\caption{Number of training and development examples by dialect in the MADAR and PADIC datasets.}
\label{tab:madar-padic-statistics}
\end{table}

\subsection{Preprocessing}
\label{sec:Preprocessing}

For the unconstrained track, English--dialect parallel data is constructed from MADAR and PADIC by translating their MSA side to English with \texttt{facebook/nllb-200-3.3B} \cite{nllb-2022-scaling} (beam size 4, max length 256). 

MADAR city-level labels are mapped to Alexandria country-level categories (e.g., Cairo/Alexandria/Aswan $\to$ Egyptian, Rabat/Fes $\to$ Moroccan). From PADIC, three dialects were incorporated, filtering out rows unaligned with MSA. The MADAR development splits and a 10\% held-out portion of PADIC were used to monitor the pretraining stage.

\subsection{Training Configuration}

Maximum sequence length is 1{,}024. Training is conducted for one epoch on a single Tesla T4 (batch size 8, gradient accumulation 4, learning rate $2\times10^{-4}$, cosine schedule, warmup 0.03, AdamW 8-bit \cite{loshchilov-hutter-2019-decoupled, dettmers-etal-2022-8bit}). Loss is computed only on the assistant completion. Training was implemented with Unsloth and the PEFT library under 4-bit NF4 quantization (bitsandbytes). To mitigate exposure bias \cite{ranzato-etal-2016-sequence}, history noising is applied: with probability $p_{\mathrm{trunc}}=0.15$ the history is truncated to a random prefix, and with probability $p_{\mathrm{noise}}=0.25$ individual previous translations undergo word-level drop/swap/duplication. The clean reference remains the training target. History noising was applied only during task fine-tuning, not during the MADAR/PADIC pretraining stage.

\subsection{Decoding and Context Handling}
\label{sec:decoding-context}

At inference time, deterministic beam-search decoding is employed with five
beams and a length penalty of 0.7 \cite{wu-etal-2016-googles}. Sampling is disabled and early stopping is enabled. The same decoding configuration is used for both the constrained and unconstrained systems. 

For dialogue translation, previously generated translations are incorporated
into the prompt as conversation history when translating subsequent turns
within the same dialogue.

\section{Results}

\paragraph{Official results.} Table~\ref{tab:dialect-results} reports per-dialect spBLEU \cite{goyal-etal-2022-flores} (SentencePiece with the flores200 tokenizer) and chrF++ \cite{popovic-2017-chrf} scores for our submitted systems on the official test set, for both the constrained and unconstrained tracks. The overall average spBLEU / chrF++ is 26.10 / 41.79 for the constrained track (4th place) and 25.09 / 41.02 for the unconstrained track (5th place). All scores above are official test-set evaluation results from the submitted systems.

\begin{table}[t]
\centering
\small
\setlength{\tabcolsep}{4pt}
\begin{tabular}{lrrrr}
\hline
& \multicolumn{2}{c}{\textbf{Constrained}} & \multicolumn{2}{c}{\textbf{Unconstrained}} \\
\textbf{Dialect} & spBLEU & chrF++ & spBLEU & chrF++ \\
\hline
Egyptian (EG)       & 30.85 & 45.43 & 30.41 & 44.89 \\
Jordanian (JO)      & 33.39 & 48.45 & 32.26 & 47.57 \\
Lebanese (LB)       & 29.70 & 44.53 & 28.09 & 43.22 \\
Libyan (LY)         & 19.82 & 37.74 & 20.14 & 37.71 \\
Moroccan (MA)       & 21.42 & 36.66 & 21.84 & 37.52 \\
Mauritanian (MR)    & 12.13 & 29.23 & 10.83 & 28.16 \\
Omani (OM)          & 26.07 & 42.06 & 24.78 & 41.59 \\
Palestinian (PS)    & 30.55 & 45.58 & 29.56 & 44.60 \\
Saudi (SA)          & 30.56 & 46.20 & 29.58 & 45.36 \\
Sudanese (SD)       & 21.59 & 37.40 & 20.49 & 36.77 \\
Syrian (SY)         & 34.73 & 50.35 & 33.64 & 49.25 \\
Tunisian (TN)       & 26.12 & 40.55 & 24.50 & 38.98 \\
Yemeni (YE)         & 22.36 & 39.05 & 20.11 & 37.64 \\
\hline
\textbf{Average}    & \textbf{26.10} & \textbf{41.79} & \textbf{25.09} & \textbf{41.02} \\
\hline
\end{tabular}
\caption{Official test-set performance (spBLEU and chrF++) across dialects for the constrained and unconstrained tracks.}
\label{tab:dialect-results}
\end{table}

\paragraph{Analysis by dialect.} Performance varies substantially across dialects in both tracks. Syrian, Jordanian, Palestinian, Saudi, Lebanese, and Egyptian consistently achieve the highest scores (all above 29 spBLEU in the constrained track), while Mauritanian is the clear outlier at the low end (12.13 constrained, 10.83 unconstrained) — roughly a third of the score of the best-performing dialects. Libyan, Moroccan, and Sudanese also trail the mid-to-high performing group, consistent with our expectation that Maghrebi and near-Maghrebi dialects, along with Mauritanian, are comparatively low-resource
\cite{harrat-etal-2020-language} and linguistically more distant from the
dialects that dominate available pretraining and task data
\cite{zaidan-callison-burch-2014-arabic}.

\paragraph{Constrained vs.\ unconstrained comparison.} Counterintuitively, the unconstrained track — where the adapter is additionally pretrained on MADAR and PADIC before task fine-tuning — scores lower on average than the constrained track (25.09 vs.\ 26.10), and underperforms on 11 of the 13 dialects. The two exceptions are Libyan (+0.32) and Moroccan (+0.42), both Maghrebi varieties for which MADAR and PADIC provide comparatively strong coverage (Table~\ref{tab:madar-padic-statistics}). This is consistent with pretraining helping specifically where it adds the most relevant, dialect-matched data, while diluting or interfering with the adapter's fit elsewhere.

However, this explanation is not fully sufficient: Sudanese receives a comparable injection of previously-absent training signal (1,600 MADAR examples, versus zero in-task training examples in the constrained setting) but \emph{degrades} under the unconstrained setup (-1.10), the opposite direction from Libyan despite an analogous zero-resource starting point. Sudanese is not a Maghrebi variety, so data availability alone does not explain the divergence. A plausible contributing factor is variation in back-translation quality across dialect families — our MADAR/PADIC pairs are constructed via NLLB-200 MSA$\to$English translation (\S\ref{sec:Preprocessing}), and translation noise on the MSA source side may be unevenly distributed across dialects, disproportionately harming some target varieties. We treat the overall pattern as a genuine, only partially understood instance of negative transfer rather than a clean Maghrebi-coverage story, and leave a controlled study of pretraining-data quality per dialect to future work.

\vspace{\baselineskip}
\paragraph{Qualitative error analysis.} To complement the aggregate metrics, we manually inspected the 20 lowest-scoring Egyptian Arabic (EG) sentence-level spBLEU outputs on the dev set. Two recurring patterns emerge. First, the model tends to translate every source word even when the natural dialectal rendering is more concise, producing fluent but overly literal output. For example, given the source:\\
``\textit{Not at all, please, go ahead. I hope everything is okay.}'', the model produces:\\
\begin{Arabic}لا خالص، تفضلي. عايزة أعرف كل حاجة تمام؟\end{Arabic}\\
where the reference instead condenses the turn to:\\
\begin{Arabic}اتفضل—خير؟\end{Arabic}\\
reflecting how Egyptian speakers compress routine social exchanges rather than translating each clause.

Second, the model under-generates the English code-switching that is pervasive in naturalistic Egyptian dialogue (e.g.,\ technical or borrowed terms like \textit{scan} or \textit{notification}), instead rendering them in Arabic script even when the reference preserves the Latin form: for 
``\textit{It's all scanned and confirmed in the system. You should get a notification now.}'',
the model outputs:\\
\begin{Arabic}كله اتمسح واتأكد في السيستم، المفروض يوصلك إشعار حالا.\end{Arabic}\\
against a reference of:\\
\begin{Arabic}كله اتعملة \textenglish{scan} واتأكد في السيستم. المفروض يوصلك \textenglish{notification} دلوقتي.\end{Arabic}

To quantify this, we flagged all dev examples where the reference contains Latin-script tokens but the model's prediction does not. This under-generation pattern is widespread and dialect-dependent, affecting 319 Moroccan, 148 Lebanese, 123 Tunisian, and 70 Egyptian examples, among others. The reverse case --- Latin script appearing in the prediction but absent from the reference --- is comparatively rare except for Tunisian (212 cases), suggesting the model has a general bias toward suppressing code-switching that it must actively overcome, with dialect-specific variation in how often this succeeds.

\section{Conclusion}

This work presented Rosetta, a LoRA-adapted NileChat system for context-aware English-to-dialectal-Arabic dialogue translation across thirteen regional varieties. Structured prompting that explicitly incorporates dialogue history and speaker metadata, together with history noising, enables competitive performance under tight parameter and data constraints. Under official evaluation, the constrained system ranked 4th (spBLEU 26.10), while the unconstrained configuration placed 5th (spBLEU 25.09). External sentence-level dialectal pretraining did not yield overall downstream gains and induced mild negative transfer across most dialects, with improvements restricted solely to Libyan and Moroccan and no single explanatory factor fully accounting for the pattern.


\bibliography{custom}

\clearpage
\appendix

\section{Prompt Templates}
\label{sec:appendix-prompts}

This section details the prompt templates used across our experiments. Section~\ref{sec:app-pretraining-prompt} shows the single-turn prompt template used for MADAR and PADIC adapter pretraining, and Section~\ref{sec:app-alexandria-prompt} illustrates the full multi-turn conversational prompt used for the Alexandria dataset.

\subsection{Pretraining Prompt Template (MADAR / PADIC)}
\label{sec:app-pretraining-prompt}

The pretraining prompt contains only sentence-level translation instructions along with the target country metadata:

\begin{promptbox}
\textbf{\textsf{System Prompt:}} \\
You are an expert translator. \\
- Return only the translated text. \\
- Do not add any code, explanations, comments, or any other extra text. \\
- Consider the country in your translation. 
\\[2ex]

\textbf{\textsf{User Prompt:}} \\
Translate the English sentence into Egyptian Arabic. 
\\[2ex]

\textbf{\textsf{Metadata:}} \\
- Country: EG 
\\[2ex]

\textbf{\textsf{Sentence to Translate:}} \\
It's at the end of the hall. I'll get you some right now. If you need anything else, just let me know. 
\\[2ex]

\textbf{\textsf{Assistant Response:}} \\
\textnormal{\begin{Arabic}هو في اخر القاعة . أنا حأجيبلك شويه دلوقتي . لو محتاج حاجة تانية، قولي.\end{Arabic}}
\end{promptbox}

\newpage
\subsection{Alexandria Conversational Prompt Template}
\label{sec:app-alexandria-prompt}

The task fine-tuning and inference prompt includes conversational history, participants, domain, and gender direction:

\begin{promptbox}
\textbf{\textsf{System Prompt:}} \\
You are an expert translator. \\
- Return only the translated text. \\
- Do not add any code, explanations, comments, or any other extra text. \\
- Keep the meaning and tone and respect the gender direction. \\
- Consider the country, the domain, the participants, and the speaker in your translation.
\\[2ex]

\textbf{\textsf{User Prompt:}} \\
Translate the English sentence into Egyptian Arabic (Cairene) Dialect.
\\[2ex]

\textbf{\textsf{Metadata:}} \\
- Country: EG \\
- Domain: Agriculture and farming \\
- Participants: Wholesale Buyer, Wholesale Seller \\
- Speaker: Wholesale Seller \\
- Speaker Direction: female $\rightarrow$ male
\\[2ex]

\textbf{\textsf{Conversation History:}} \\[0.8ex]
\textbf{Wholesale Buyer:} Good morning. I'm looking to source 10 tons of premium artichokes for export. People say the best quality in Obour comes from your section, is that right?
\\[1.2ex]
\textbf{Translation:} \\[0.8ex]
\textnormal{\begin{Arabic}صباح الخير، عايز عشرة طن من الخرشوف الكويس للتصدير، بيقولوا ان احسن جودة في سوق العبور بتيجي من عندكم، صحيح؟\end{Arabic}}
\\[2ex]

\textbf{\textsf{Sentence to Translate:}} \\[0.8ex]
Good morning to you. You heard correctly. My artichokes are the best you'll find. They are top-grade, perfect for export. Let me show you a sample.
\\[2ex]

\textbf{\textsf{Assistant Response:}} \\[0.8ex]
\textnormal{\begin{Arabic}صباح النور، سمعك مظبوط، الخرشوف بتاعي من أحسن اللي موجود في السوق، من أحسن الأنواع، معمول للتصدير، خليني أوريك عينة.\end{Arabic}}
\end{promptbox}

\end{document}